\documentclass[10pt,twocolumn,letterpaper]{article}

\usepackage{cvpr}
\usepackage{times}
\usepackage{epsfig}
\usepackage{graphicx}
\usepackage{amsmath}
\usepackage{amssymb}
\usepackage {multirow}
\usepackage{subfigure}
\usepackage{booktabs}
\usepackage{enumitem}

\def\eg{\textit{e.g.}}
\def\ie{\textit{i.e.}}

\def\etal{\textit{et al. }}

\usepackage[breaklinks=true,bookmarks=false]{hyperref}

\cvprfinalcopy 

\def\cvprPaperID{2252} 

\begin{document}

\title{Learning Cross-Modal Deep Representations for 
Robust Pedestrian Detection}

\author{Dan Xu$^1$, Wanli Ouyang$^{2,3}$, Elisa Ricci$^{4,5}$, Xiaogang Wang$^2$, Nicu Sebe$^1$\\
$^1$University of Trento, $^{2}$The Chinese University of Hong Kong\\
$^3$The University of Sydney, $^4$Fondazione Bruno Kessler, $^5$University of Perugia\\
{\tt\small \{dan.xu, niculae.sebe\}@unitn.it, eliricci@fbk.eu, \{wlouyang, xgwang\}@ee.cuhk.edu.hk} 
}

\maketitle

\begin{abstract}
This paper presents 
a novel method for detecting pedestrians under adverse illumination
conditions. Our approach relies on a novel cross-modality learning
framework and it is based on two main phases. First, given a multimodal dataset, a deep convolutional
network is employed to learn a non-linear mapping, modeling the relations between RGB and thermal data.
Then, the learned feature representations are transferred 
to a second deep network, which receives as input an RGB image
and outputs the detection results. In this way, features
which are both discriminative and robust to bad illumination conditions are learned. Importantly, at test time, only the second pipeline is considered and no thermal data are required. 
Our extensive evaluation demonstrates that the proposed approach outperforms the
state-of-the-art on the challenging KAIST multispectral pedestrian dataset
and it is competitive with previous methods on the popular Caltech dataset.

\end{abstract}

\section{Introduction}
\label{intro}



Great strides in pedestrian detection research \cite{benenson2014ten} have been made 
for challenging situations, such as cluttered background, substantial occlusions and tiny target appearance. 
As for many other computer vision tasks, in the last few years
significant performance gains have been achieved thanks to approaches based on 
deep networks \cite{ouyang2013modeling,angelovareal,li2015scale,tian2015pedestrian}.
Additionally, the adoption 
of novel sensors, \eg \ thermal and depth cameras, has provided new opportunities, advancing the state-of-the-art on pedestrian
detection by tackling problems such as adverse illumination conditions and occlusions \cite{hwang2015multispectral,gonzalez2016pedestrian,premebida2014pedestrian}.
However, the vast majority of wide camera networks in surveillance systems still employ traditional RGB sensors and
detecting pedestrians in case of illumination
variation, shadows, and low external light is still a challenging open issue.


\begin{figure}[t]
\centering
\includegraphics[width=0.96\linewidth]{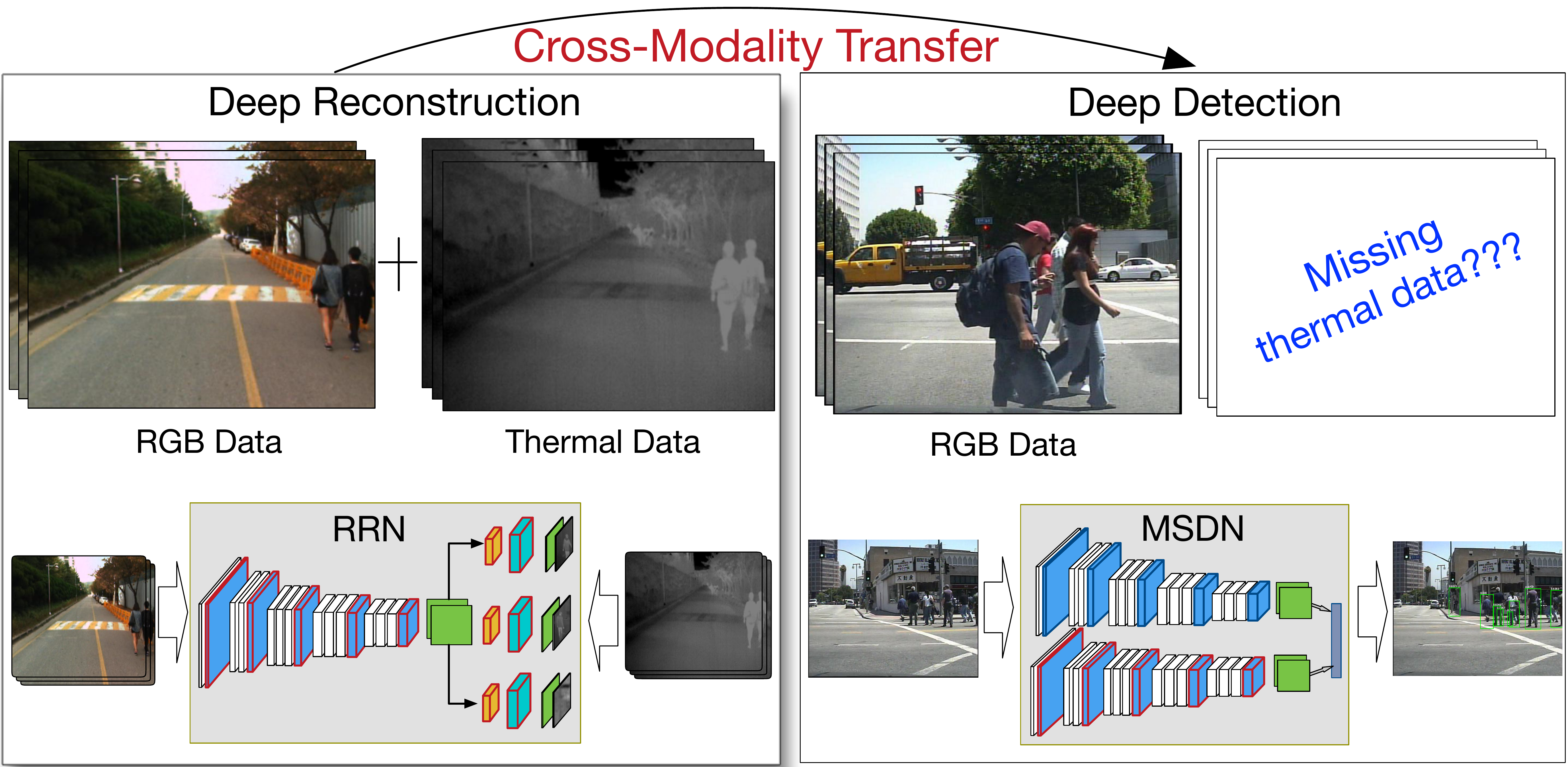}
\caption{Overview of our framework.
Our approach relies on two networks. The first network, named Region Reconstruction Network
(RRN) is used to learn a non-linear feature mapping between RGB and thermal image pairs. Then, the learned model 
is transferred to a target domain where thermal inputs are no longer available and 
a second network, the Multi-Scale Detection Network (MDN), is used for learning an RGB-based pedestrian detector.
}
\vspace{-0.4cm}
\label{fig:idea3}
\end{figure}

This paper introduces a novel approach based on Convolutional Neural Networks (CNN) to address this problem. 
Our method is inspired by recent works demonstrating that learning deep 
representations from cross-modal data is greatly beneficial for detection and recognition tasks \cite{gupta2016cross,hoffman2016}. 
However, most approaches assume the availability of large annotated datasets. In the specific case of pedestrian
detection, the community can rely on a great abundance of visual data gathered with surveillance cameras, cars and robotic platforms, but
there are few labeled multi-modal datasets. Therefore, motivated by the successes of recent 
unsupervised deep learning techniques, we introduce an approach for learning cross-modal 
representations for pedestrian detection which does not require pedestrian bounding box annotations.
More specifically,
we propose leveraging information 
from multispectral data and using a deep convolutional network to learn a non-linear mapping from RGB to thermal images without human supervision.
This cross-modal mapping is then exploited by integrating the learned representations into a second deep architecture,
operating on RGB data and effectively modeling multi-scale information. 
Importantly, at test time, thermal data are not needed and
pedestrian detection is performed only on color images. 

\begin{figure}[t]
\centering
\includegraphics[width=0.98\linewidth, height=1.5in]{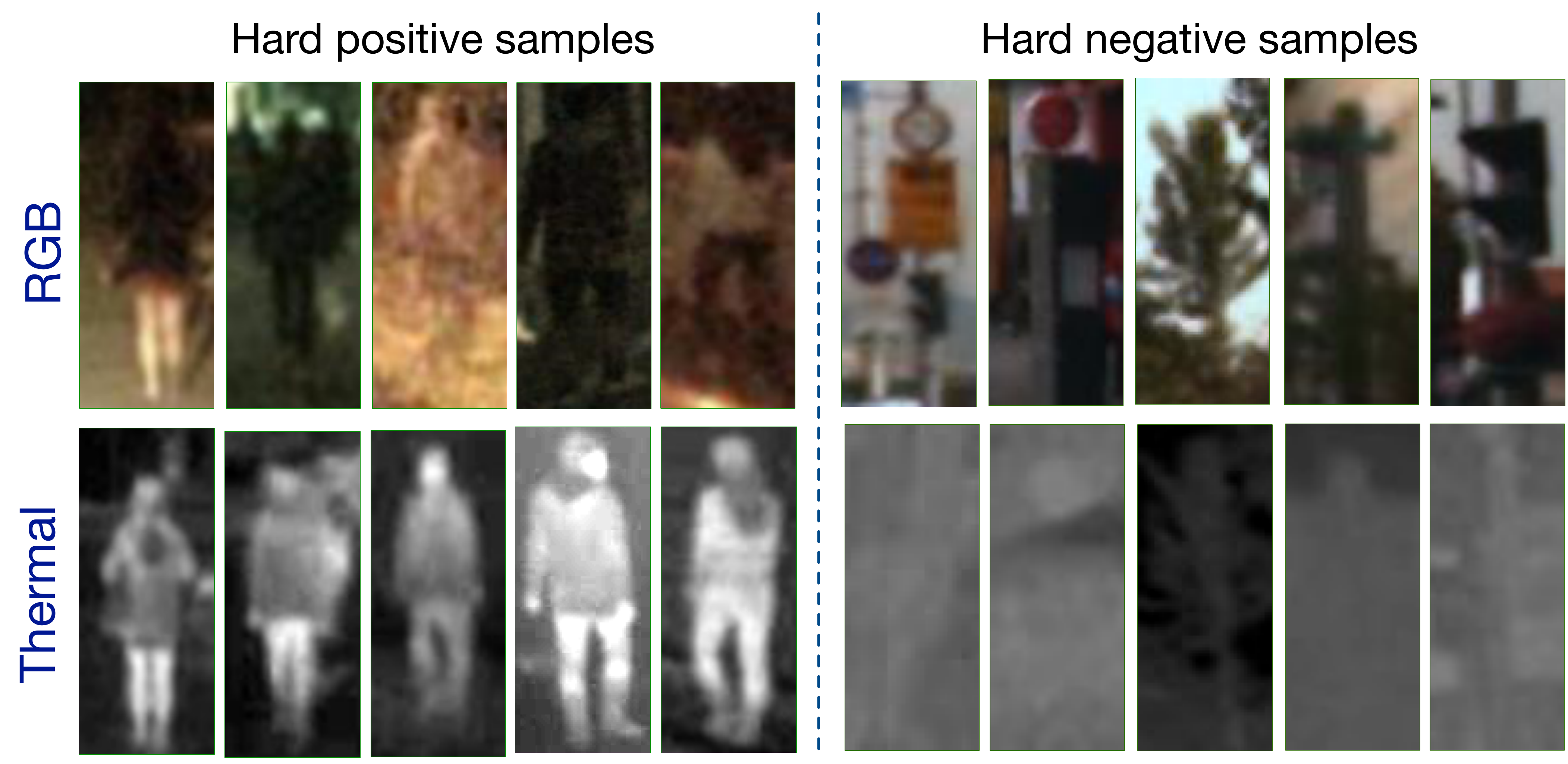}
\caption{Motivation of this work. By exploiting thermal data in addition to RGB samples, it is easier to 
discriminate among pedestrians and background clutter. 
}
\vspace{-0.5cm}
\label{fig:idea2}
\end{figure}

Figure \ref{fig:idea3} depicts an overview of the proposed approach. 
Our intuition, illustrated in Fig.\ref{fig:idea2}, is that, 
by exploiting multispectral data with the proposed method, 
it is easier to distinguish hard negative samples in color images (\eg, electric poles or trees with 
appearance similar to pedestrians), thus improving the detection accuracy. 
Experimental results on publicly available datasets, where several frames are
captured under bad illumination conditions, demonstrate the advantages of our approach over previous methods. To summarize the main contributions of this work are:
\setlist[itemize]{leftmargin=*,label=\scalebox{.8}{\textbullet}}
\noindent\begin{itemize} \vspace{-0.2cm}
 \item We introduce a novel approach for learning and transferring cross-modal feature representations
for pedestrian detection. With the proposed framework, data from the auxiliary modality (\ie \ thermal data) are used 
as a form of supervision for learning CNN features from RGB images. There are two fundamental advantages in our strategy. First,
multispectral data are not employed at the test phase. This is crucial when deploying robotics and surveillance systems, 
as only traditional cameras are needed, significantly decreasing costs. Second, no pedestrian annotations are required in the thermal domain. 
This greatly reduces human labeling efforts and permits to exploit large data collections of RGB-thermal image pairs.
\item To our knowledge, this is the first work specifically addressing the problem of pedestrian detection under adverse illumination conditions
with convolutional neural networks. Previous works mostly adopted hand-crafted descriptors and integrated the thermal modality
by using additional input features \cite{hwang2015multispectral,socarras2011adapting}. Our approach is based on two novel
deep network architectures, specifically designed for unsupervised cross-modal feature learning and for effectively
transferring the learned representations.\vspace{-0.2cm}
\item Through an extensive experimental evaluation, we demonstrate that our framework outperforms the state-of-the-art on the novel 
KAIST multispectral pedestrian dataset 
\cite{hwang2015multispectral} and it is competitive with previous methods on the popular Caltech dataset \cite{dollar2009pedestrian}.
\end{itemize}

This paper is organized as follows. Section \ref{related} outlines
related work on pedestrian detection and cross-modal feature learning. Section \ref{method} describes the proposed
framework for learning features robust to illumination variations in the context of pedestrian detection. 
Experimental results to demonstrate the benefits of 
our approach are presented in
Section \ref{results}. We conclude with key remarks in Section \ref{conclusions}.

\begin{figure*}[t]
\centering
\includegraphics[width=0.86\textwidth]{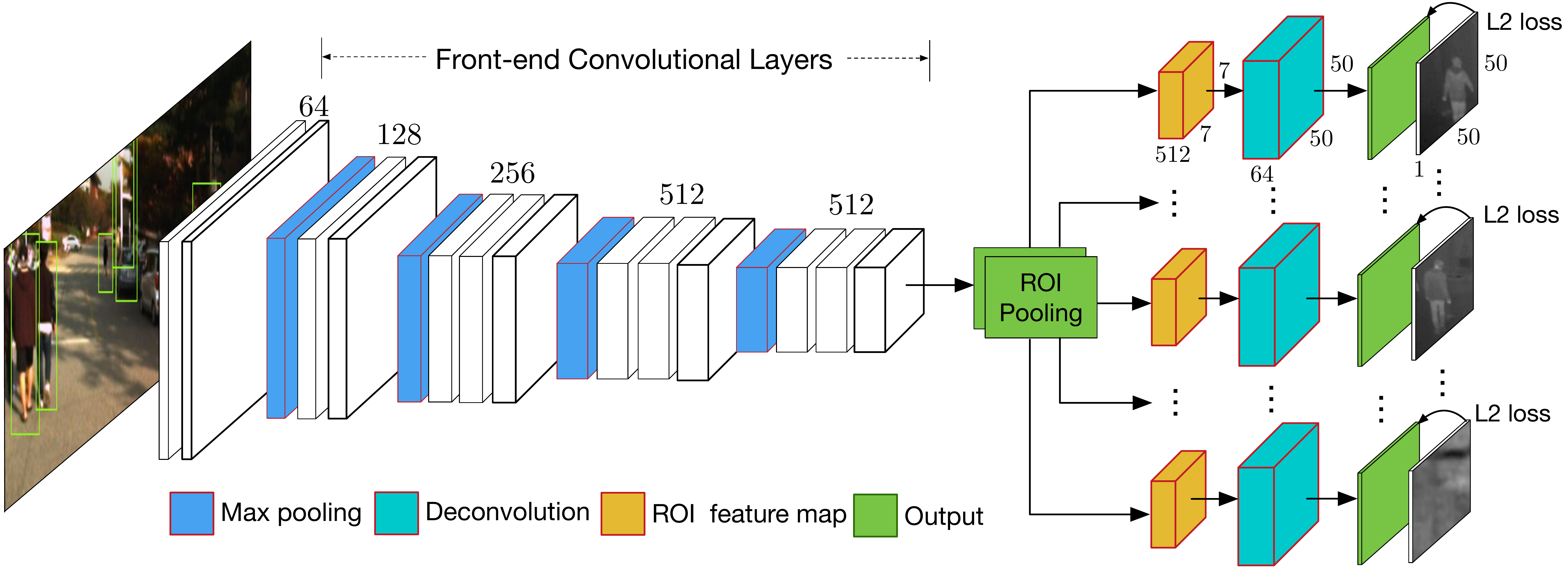}
\caption{Architecture of the Region Reconstruction Network: a deep convolutional network trained
for reconstructing thermal images from the associated RGB data. Best viewed in color.}
\label{fig:encdec}
\vspace{-0.4cm}
\end{figure*}



\section{Related Work}
\label{related}
Research topics closely related to this work are pedestrian detection from surveillance videos and deep learning approaches operating on multimodal data. 
Below, we present a review of the most recent works on these topics.

\textbf{Pedestrian Detection.} Due to its relevance in many fields, such as robotics and video surveillance, the problem of
pedestrian detection has received considerable interests in the research community. Over the years, a large variety of features and algorithms
have been proposed for improving detection systems, both with respect to speed \cite{viola2005detecting,benenson2012pedestrian,angelovareal,li2015scale} and accuracy 
\cite{yan2013robust,paisitkriangkrai2014strengthening,zhang2014informed,zhang2015filtered,FRCNN,tian2015pedestrian}.

Recently, notable performance gains have been achieved with the adoption of powerful deep networks \cite{ouyang2013modeling,angelovareal}, 
thanks to their ability to learn discriminative features directly from raw pixels. In \cite{sermanet2013pedestrian}, a CNN pre-trained
with an unsupervised method based on convolutional sparse coding 
was presented. The occlusion problem was addressed in \cite{ouyang2012discriminative}, where a deep belief net was employed to learn the 
visibility masks for different body parts. This work was extended in \cite{ouyang2013single} to model relations among multiple targets. 
More recently, in~\cite{tian2015deep} DeepParts, a robust framework for handling severe occlusions, was presented.
Differently from previous deep learning models addressing the occlusion problem, DeepParts does not rely on a single detector but it is based on
multiple part detectors.
Tian \etal \cite{tian2015pedestrian} learned discriminative representations for pedestrian detection by considering 
semantic attributes of people and scenes. 
Cai \etal \cite{cai2015learning} introduced Complexity-Aware Cascade Training (CompACT), successfully integrating many heterogeneous features, both hand crafted and 
derived from CNNs. Zhang \etal \cite{zhang2016faster} presented an approach based on the Region Proposal Network (RPN) \cite{ren2015faster} and boosted 
forests.

Other works focused on improving the computational times of CNN-based pedestrian detectors. 
For instance, 
Angelova \etal \cite{angelovareal} proposed the DeepCascade method, \ie \ a cascade of deep neural networks, and demonstrated a considerable gain in 
terms of detection speed. An in-depth analysis of different deep networks architectural choices for pedestrian detection was provided in
\cite{hosang2015taking}.
To our knowledge, none of these previous works considers multi-modal data or tackles the problem of pedestrian detection under 
adverse illumination conditions.

Previous works have considered transferring information from other domains for constructing scene-specific pedestrian detectors. 
Wang \etal \cite{wang2012transferring} proposed an unsupervised approach where target samples are collected by exploiting contextual 
cues, such as motions and scene geometry. Then, a pedestrian detector is built by re-weighting labeled source samples, \ie \ by assigning
more importance to samples more similar to target data.
This approach was later extended in \cite{zeng2014deep} to learn deep feature representations.
Similarly, in \cite{cao2013transfer} a sample selection scheme to reduce the discrepancy between source and target distributions was presented. 
Our approach 
is substantially different, as we do not restrict our attention to adapt a generic model to a specific scene
and we tackle the problem of 
transferring knowledge among different modalities. 

\textbf{Learning Cross-modal Deep Representations.} In the last few years deep networks have been successfully 
applied to learning feature representations from multi-modal data \cite{karpathy2014deep,xu2017detecting,xu2015learning}.
However, the problem of both learning and transferring cross-modal features has been rarely investigated. Notable exceptions are the works in 
\cite{christoudias2010learning,srivastava2012multimodal,socher2013zero,gupta2016cross,hoffman2016}. 
Among these, the most similar to ours are \cite{christoudias2010learning,srivastava2012multimodal,hoffman2016}. In 
\cite{christoudias2010learning,srivastava2012multimodal} the idea of 
hallucinating data from other modalities 
was also exploited. However, our CNN-based approach is substantially different, 
since the work in \cite{srivastava2012multimodal} considered
Deep Boltzmann Machines, while in \cite{christoudias2010learning} the mapping between different modalities was learned with Gaussian Processes.
In \cite{hoffman2016} the problem of object detection from RGB data was addressed and depth images were used as additional information
available only at training time. Similarly to \cite{hoffman2016}, our detection network simultaneously use cross-modal features learned from a 
source domain and representations specific of the target scenario. However, in \cite{hoffman2016} 
labeled data were available in the original domain. Oppositely, in our framework we learn cross-modal features in an unsupervised
setting, \ie \ we do not require any annotation in the thermal domain. In this way, it is possible to exploit huge multispectral datasets.



\begin{figure*}[t]
\centering
\includegraphics[width=0.86\textwidth]{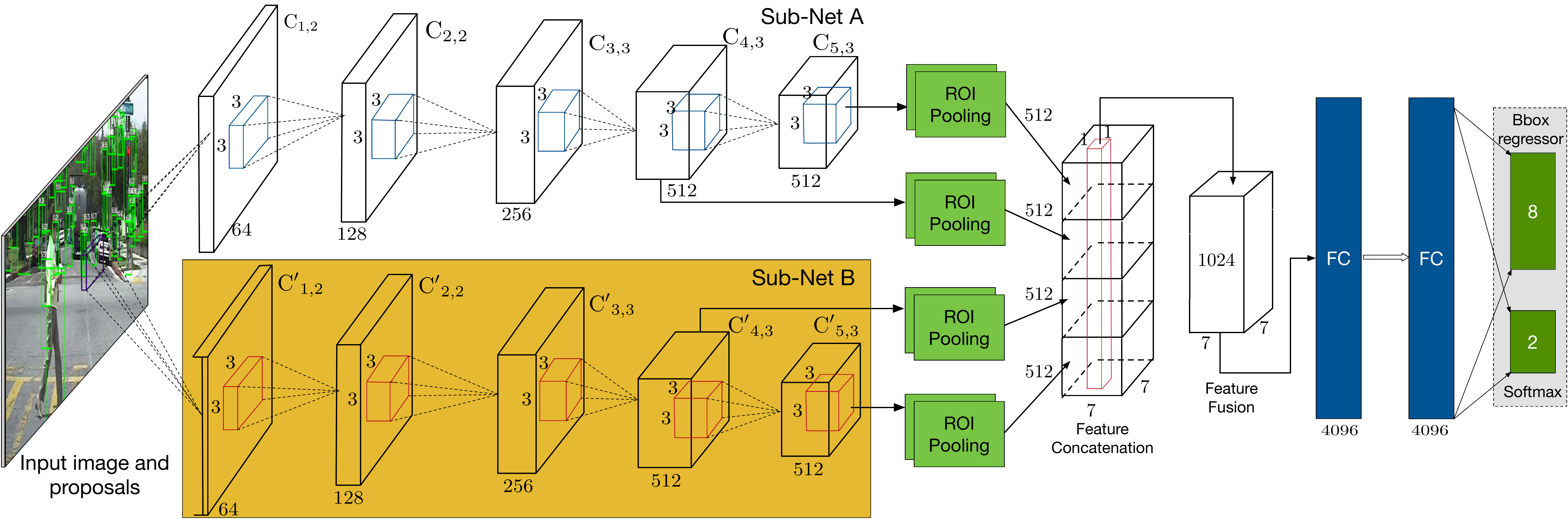} 
 \caption{Architecture of the Multi-Scale Detection Network. Two sub-networks (Sub-Net A and Sub-Net B) with the same structure are used in MSDN. 
 The parameters of all the convolutional layers of Sub-Net B (highlighted in yellow) are transferred
 from the Region Reconstruction Network. 
 }
 \label{fig:cnn-cmt}
 \vspace{-0.4cm}
 \end{figure*}
 
\section{Learning and transferring cross-modal deep representations}
\label{method}
In this section we present the proposed framework. We first provide an overview of our approach and we describe in details  
the CNN architectures we design to reconstruct thermal data from RGB input and to transfer the learned cross-modal representations 
for the purpose of robust pedestrian detection.

\subsection{Overview}
As outlined in Section \ref{intro}, the proposed framework (Fig.\ref{fig:idea3}) in based on two different convolutional neural networks, 
associated to the
reconstruction and to the detection tasks, respectively. 
The first deep model, \ie \ the Region Reconstruction Network (RRN), is a fully convolutional network 
trained on pedestrian proposals collected from RGB-thermal image pairs in an unsupervised manner. 
RRN is used to learn a non-linear mapping from the RGB channels to the thermal channel. 
In the target domain only RGB data are available and a second deep network, the Multi-Scale Detection Network (MSDN), 
embedding the parameters transferred from RRN, is used for robust pedestrian detection.
MSDN takes a whole RGB image and a number of pedestrian proposals as input and outputs the detected bounding boxes with associated scores. 
In the test phase, detection is performed with MSDN and only RGB inputs are needed. In the following we describe the details of the proposed deepnet framework.

\subsection{Region Reconstruction Network} 

The aim of RRN is to reconstruct thermal data from the associated RGB images.
The design of the RRN architecture is driven by two main needs. First, in order to avoid human annotation efforts, thermal information
should be recovered with an unsupervised approach. While our approach uses the thermal image as deep supervision for the reconstruction task, it essentially requires only very weak supervision information (\textit{i.e.}, the pair-wise information). However, in the RGB-T data collection phase, we easily obtain the pair-wise information. The most expensive part in terms of human effort is to annotate the pedestrian bounding boxes. The proposed approach does not require these extra human-annotations. Second, as multispectral data are expected to be especially useful 
for hard positive and negative samples (Fig.\ref{fig:idea2}), instead of attempting to reconstruct the entire thermal images, it is more appropriate to specifically focus on bounding boxes which are likely to contain pedestrians. Therefore, in this paper we propose to exploit a pretrained generic pedestrian detector (\eg~ACF \cite{dollar2014fast}) to extract a set of pedestrian proposals (containing true positives and false positives) from RGB data and design a deep model which reconstructs the associated thermal information. 

\par The proposed RRN network is illustrated in Fig.\ref{fig:encdec}. The input of RRN is a three-channel RGB image and a set of 
associated pedestrian proposals. RRN consists of a front-end convolutional subnetwork and a back-end reconstruction subnetwork. 
Although in our implementation the front-end convolutional layers exploit the VGG-13 network structure~\cite{simonyan2014very}, RRN alternatively 
supports other architectures. After the last convolutional layer of the front-end subnetwork, an ROI pooling layer~\cite{FRCNN} is added.
For each ROI, feature maps with size $512 \times 7 \times 7$ are generated. Considering the small size of the ROI feature maps, in order to 
effectively reconstruct the regions of thermal images associated to pedestrians, we apply a deconvolutional layer to upsample 
the ROI feature maps (output size $50 \times 50$) and reduce the number of output channels to $64$  
to ensure smooth convergence during training. Different from many previous works (\eg~\cite{xie2015holistically}) which simply consider a bilinear upsampling operator, in the 
deconvolutional layer we learn the upsampling kernels (kernel size $4$, stride 8 and pad 1). 
After the deconvolutional layer, a Rectified Linear Unit (ReLU) layer is applied. Then, reconstruction maps corresponding to each proposal are generated using a convolutional layer (kernel size 3, pad 1). Finally,
a square loss is considered to compute each reconstruction map and the whole network is optimized with
back-propagation.
\par In the widely used Fast- or Faster-RCNN frameworks, the groundtruth pedestrian bounding boxes are used to determine the ratio of true positive and false positive samples, and then construct fixed-size training mini-batches. To avoid using the carefully annotated groundtruth bounding boxes, we construct each training mini-batch using pedestrian proposals generated by thresholded generic ACF from one randomly selected training image, since the number of the proposals corresponding to each training image dynamically changes, our approach thus implements a dynamic mini-batch size during training.
\subsection{Multi-Scale Detection Network}
MSDN is specifically designed to perform pedestrian detection from RGB images by exploiting 
the cross-modal representations learned with RRN. Inspired by previous works demonstrating the
importance of considering multi-scale information in detection tasks~\cite{zhang2016faster, xu2017learning},
we introduce a detection network which fuses multiple feature maps derived from 
ROI pooling layers.  

MSDN architecture seamlessly integrates two sub-networks (Sub-Net A and Sub-Net B), 
as illustrated in Fig.~\ref{fig:cnn-cmt}. 
Sub-Net A has 13 convolutional layers, organized in five blocks. As depicted in Fig.\ref{fig:cnn-cmt}, $\mathrm{C}_{m, n}$ denotes the $m$-th block 
with $n$ convolutional layers with the same size filters. 
Max pooling layers are added after the convolutional layers, and the ReLU non-linearity is applied to the output of each convolutional layer. 
An RoI (Region of Interest) pooling layer~\cite{FRCNN} is applied to the last two convolutional blocks to extract  
feature maps of size $512 \times 7 \times 7$ for each pedestrian proposal. We consider these two blocks, as our experiments show that 
this strategy represents the optimal trade-off between computational complexity and accuracy. Sub-Net B has the same structure of 
Sub-Net A but, since its main goal is to transfer cross-modality mid-level representations, the parameters of the 13 convolutional 
layers ($\mathrm{C'}_{1,2}$ to $\mathrm{C'}_{5,3}$) are derived from the associated layers of RRN. Indeed,
the convolutional blocks from RRN produce a compact feature representation
which captures the complex relationship among the RGB and the thermal domain. Therefore, they are embedded
in MSDN, such as to allow the desired knowledge transfer.

The feature maps derived from the RoI pooling layers of the two sub-networks are then combined with a concatenation layer
and a further convolutional layer with 1024 channels is applied. As the size of the RoI feature maps is small, 
we set the kernel size equal to 1 in the convolutional layer. Then, 
two fully connected layers of size 4096 follow.
Finally, two sibling layers are used, one that outputs softmax probability estimates over pedestrian and 
background classes, and another that provides the associated bounding-box offset values for pedestrian localization.

\subsection{Optimization}
As discussed above, the proposed cross-modal framework is based on two different deep networks. Therefore, 
the training process also involves two main phases. 

In the first phase, RRN is trained on multispectral data. 
The front-end convolutional layers of RRN are initialized using the parameters of the 13 convolutional layers 
of the VGG-16 model \cite{simonyan2014very} pretrained on ImageNet dataset. The remaining parameters are randomly initialized. 
Stochastic Gradient Descent (SGD) is used to learn the network parameters. 
In the second phase, the parameters of MSDN are optimized using RGB data and pedestrian bounding box annotations in the target domain. 
We first train Sub-Net A by adding the common parts of MSDN (\ie~from the feature concatenation layer to the two sibling layers). In this
case the size of the feature maps in the concatenation and in the following convolutional layers
is $1024 \times 7 \times 7$ and $512 \times 7 \times 7$, 
respectively. The pretrained VGG-16 model is also utilized to initialize Sub-Net A.
The convolutional layers of Sub-Net B are initialized with the corresponding parameters of RRN. Then, fine-tuning is performed 
using the RGB data of the target domain. The whole MSDN optimization is based on back-propogation with SGD.

\subsection{Pedestrian detection}
In the detection phase, given a test RGB image, we adopt the standard protocol. First, region proposals are extracted, similarly
to the training phase.
Then, the input image and the proposals are fed into MSDN. 
The softmax layer outputs the class score and the bounding box regressor indicates the estimated image coordinates. 
To reduce the redundancy of the proposals, non-maximum suppression is employed based on the prediction score of each proposal, 
setting an intersection over union (IoU) threshold $\delta$.

\section{Experiments}
\label{results}

To evaluate the effectiveness of the proposed framework, we performed experiments on two publicly available datasets: 
the recent KAIST multispectral pedestrian dataset \cite{hwang2015multispectral} and the popular Caltech pedestrian 
dataset \cite{dollar2009pedestrian}. In the following we describe the details of our evaluation.

\subsection{Datasets}
The \textbf{KAIST} multispectral pedestrian dataset \cite{hwang2015multispectral} contains images 
captured under various traffic scenes with different illumination conditions (\ie \ data recorded both during day and night). 
The dataset consists of 95,000 aligned RGB-thermal image pairs, of which 50,200 samples are used for training and the rest for testing.
A total of 103,128 dense annotations corresponding to 1,182 unique pedestrians are available. 
We follow the protocol outlined in~\cite{hwang2015multispectral} in our experiments. The performance is evaluated on three different test sets, denoted as 
\textit{Reasonable all}, \textit{Reasonable day} and \textit{Reasonable night}. \textit{Reasonable} indicates that the pedestrians are not/partially 
occluded with more than 55 pixels height. The day and night sets are obtained from the \textit{Reasonable all} set according to the capture time.

The \textbf{Caltech} pedestrian dataset \cite{dollar2009pedestrian} consists of about 10 hours of 30Hz video collected from a 
vehicle driving through urban traffic. The dataset contains 250,000 frames with 350,000 bounding boxes manually annotated and 
associated to about 2,300 unique pedestrians. Following previous works \cite{tian2015pedestrian,li2015scale}, we strictly adopt 
the evaluation protocol in \cite{dollar2009pedestrian} measuring the log average miss rate over nine points ranging from $10^{-2}$ to $10^{0}$  
False-Positive-Per-Image (FPPI).
Our evaluation is conducted on both Caltech-All and Caltech-Reasonable settings.

Our approach uses RGB-thermal data for training, but in the test phase only requires RGB images as input. In all our experiments the KAIST 
training dataset is used to learn the RRN.  
Then, the performance of MSDN is assessed on the Caltech test set and on the RGB test frames of KAIST. The training and testing images 
of both datasets are resized (800 pixels height) to generate ROI feature maps with higher resolution useful
for our reconstruction and detection tasks.

 \begin{figure}[!t]
\centering
\includegraphics[width=0.49\textwidth]{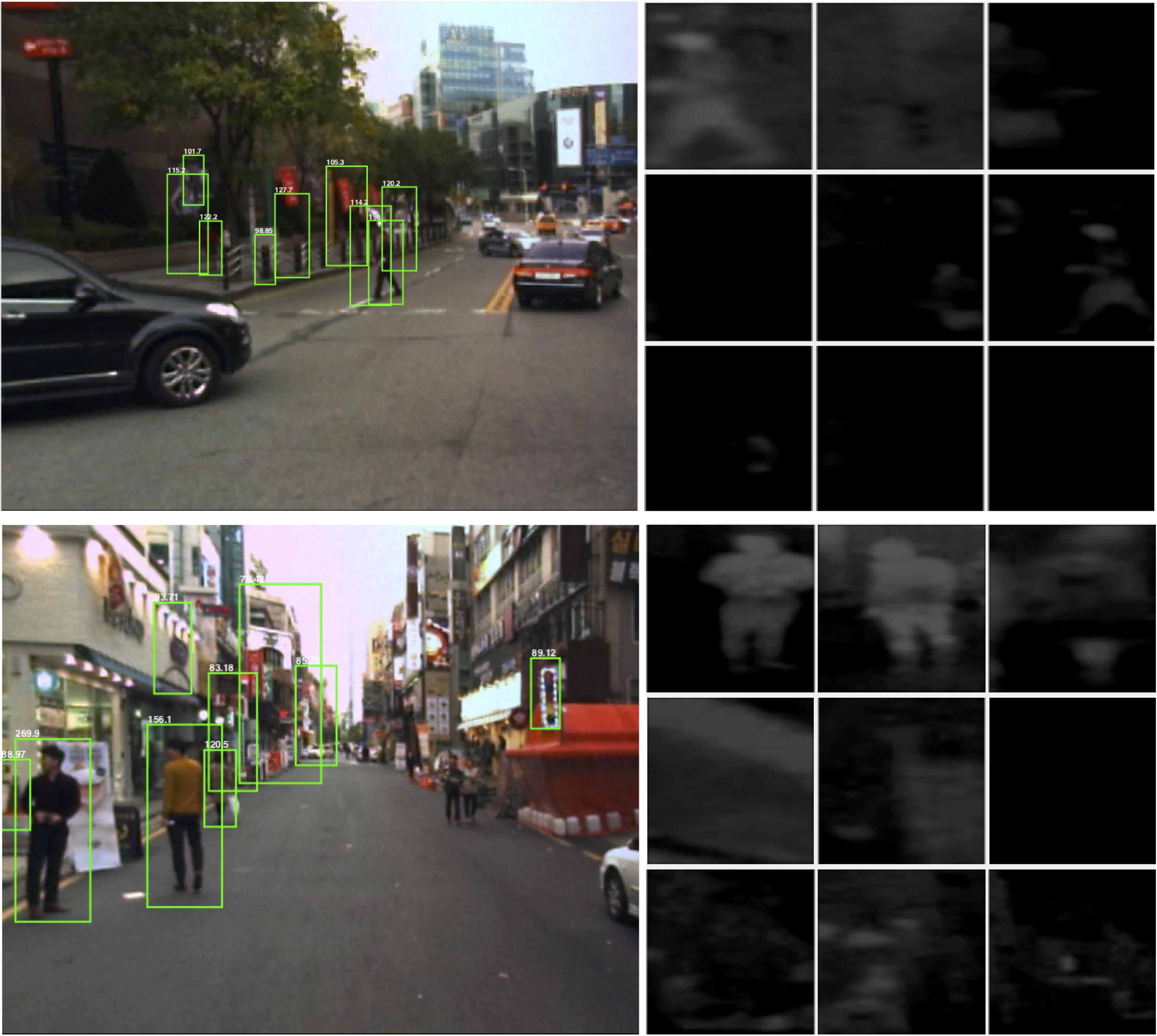} 
 \caption{KAIST dataset. Reconstructed regions of thermal images ($50\times 50$ pixels) associated to the top nine detected pedestrian windows
 from ACF. 
 }
 \label{fig:reconstruction_results}
 \vspace{-0.4cm}
 \end{figure}

\subsection{Experimental setup}
Our framework is implemented under \textit{Caffe}, and our evaluation is conducted on an Intel(R) Xeon(R) CPU E5-2630 with a 
single CPU core (2.40GHz), 64GB RAM and a NVIDIA Tesla K40 GPU. 

We employ ACF~\cite{dollar2014fast} to generate pedestrian proposals for training both the reconstruction and the detection 
network {with a low detection threshold of -70 as in~\cite{li2015scale} to obtain a high recall of pedestrian regions.} In the test phase 
we also use ACF and consider the test proposals available online\footnote{\url{http://www.vision.caltech.edu/Image$\_$Datasets/CaltechPedestrians/}}. 
It is worth nothing that, while we focus on ACF, our cross-modality learning approach can be used in combination with an arbitrary proposal method.

For training the reconstruction network, we use the whole training set of the KAIST dataset. As thermal images captured from an infrared device 
have relatively low contrast and significant noise, we perform some basic processing, such as adaptive histogram equalization and denoising. 
By computing pedestrian proposals applying ACF, we end up creating a dataset of about	20K frames for training the region reconstruction network. All the frames 
are then horizontally flipped for data augmentation. We generate mini-batch of reconstruction RoIs from randomly chosen two images, and a fixed learning rate $\lambda_r=10^{-9}$ is used to guarantee smooth convergence. We train the RRN for about 10 epochs.

For training the detection network on the the Caltech dataset we follow~\cite{zhang2015filtered} and we construct a training set where
every $3^{rd}$ frame is used. Instead, for the KAIST dataset we adopt the standard training protocol and every $20^{th}$ frame is considered. 
For both datasets, we use the same protocol for training MSDN. Similarly to RRN training, the data 
are flipped horizontally for the purpose of data augmentation. Each mini-batch consists of 128 pedestrian proposals randomly chosen from one 
training image. Positive samples with a ratio of 25\% are taken from the proposals which have an IoU overlap with the ground truth of more 
than $0.5$, while negative samples are obtained when the IoU overlap is in the range of [0, 0.5]. 
Stochastic gradient descent is used to optimize MSDN with the 
momentum and the weight decay parameters set to 0.9 and 0.0005, respectively.
The network is trained for 8 epochs using 
an initial learning rate of $0.001$ and drop by 10 times at the $5^{th}$ epoch. 


 \begin{figure}[t]
\centering
\includegraphics[width=0.48\textwidth]{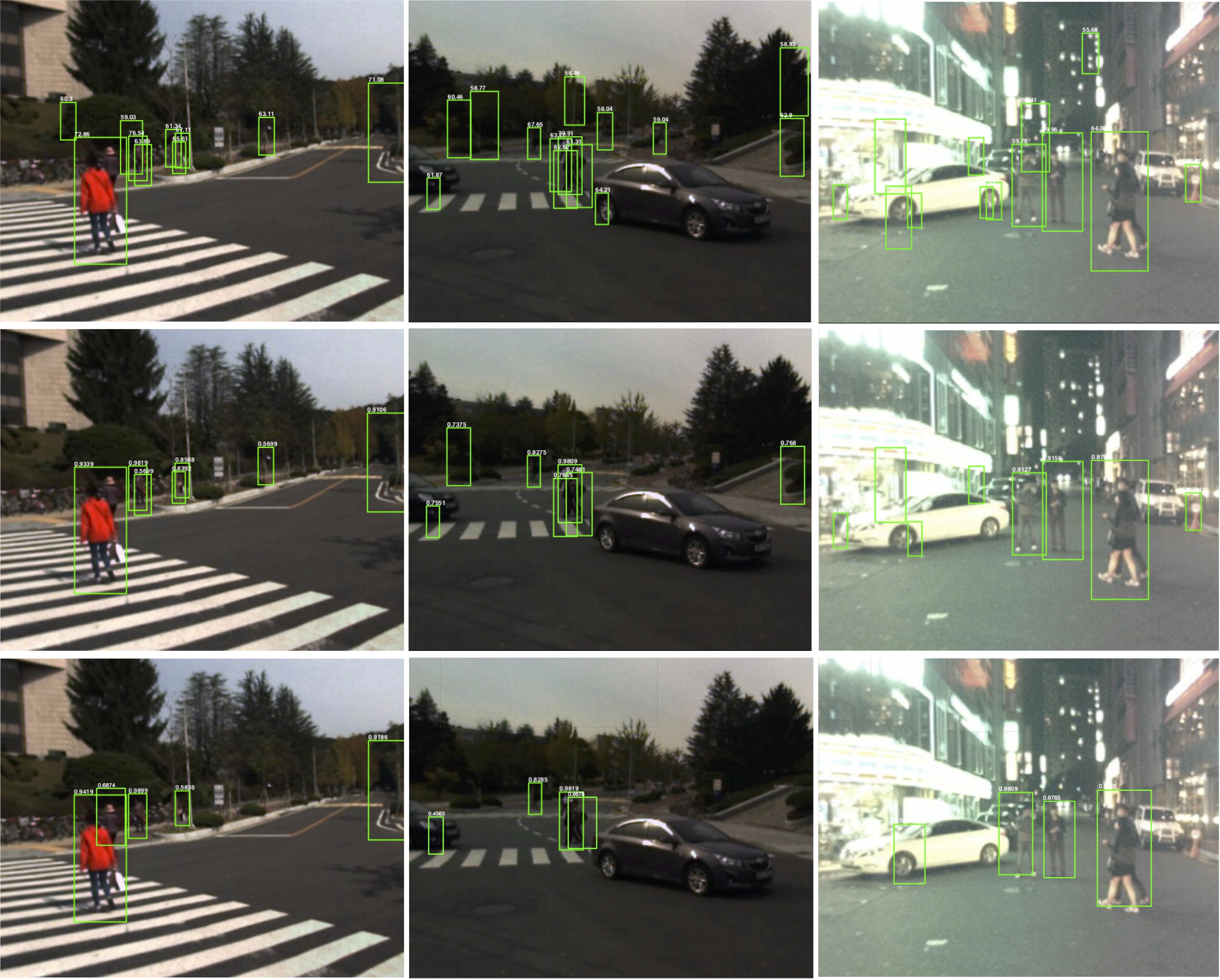} 
 \caption{Examples of pedestrian detection results under different illumination conditions on the KAIST multispectral pedestrian dataset: 
 (top) ACF detector, (middle) CMT-CNN-SA, (bottom) CMT-CNN. 
}
 \label{fig:detection_results}
 \end{figure}

 \begin{table}[t] 
\centering
\resizebox{1\linewidth}{!} {
\begin{tabular}{l|ccc}
\toprule
Methods & All & Day & Night \\\hline \hline
CMT-CNN-SA & 54.26\% & 52.44\% & 58.97\% \\
CMT-CNN-SA-SB(Random) & 56.76\% & 54.83\% & 61.24\% \\
CMT-CNN-SA-SB(ImageNet) & 52.15\% & 50.71\% & 57.65\% \\
CMT-CNN& 49.55\% & 47.30\% & 54.78\% \\
\bottomrule
\end{tabular}
}
\caption{Comparison of different methods on the KAIST multispectral datasets including 
reasonable all, reasonable day and reasonable night settings. 
}
\label{tab:cmtcmpKAIST}
\end{table}

 \begin{figure*}[t] 
\centering
\subfigure[Reasonable all]{\includegraphics[width=0.33\linewidth]{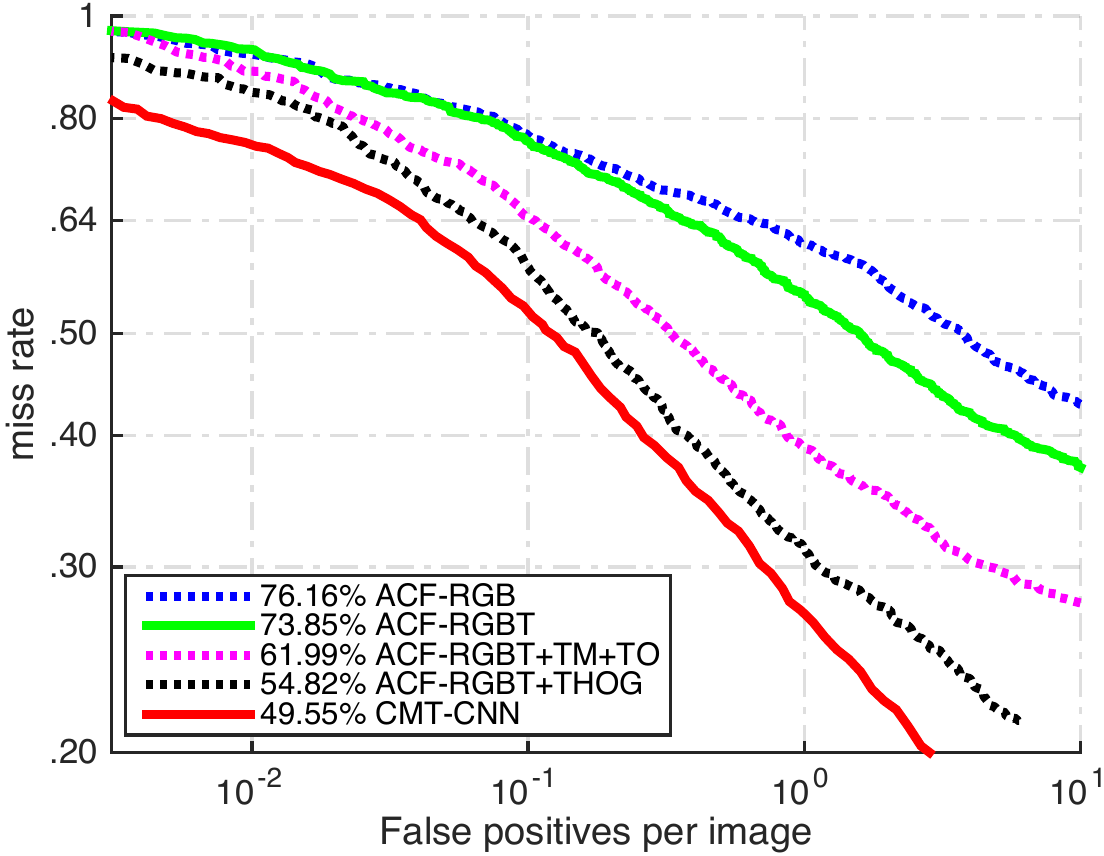}}%
\subfigure[Reasonable day]{\includegraphics [width=0.33\linewidth]{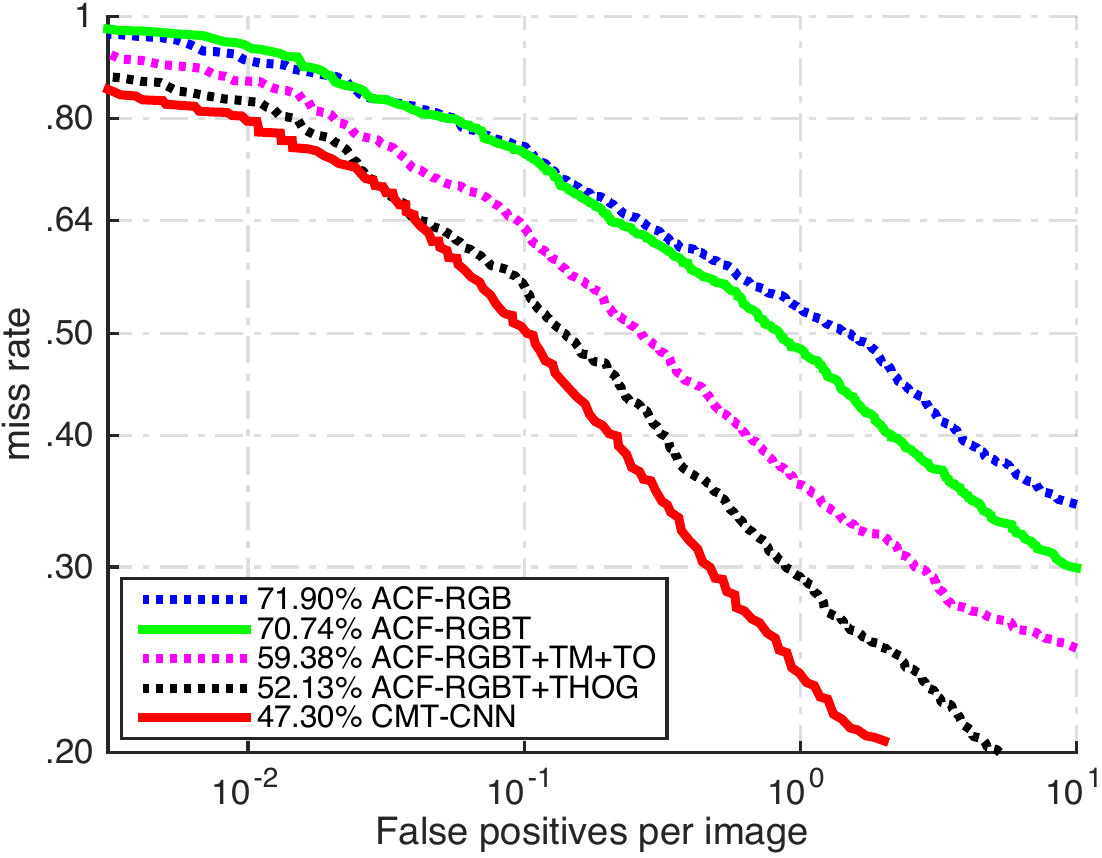}}
\subfigure[Reasonable night]{\includegraphics [width=0.33\linewidth]{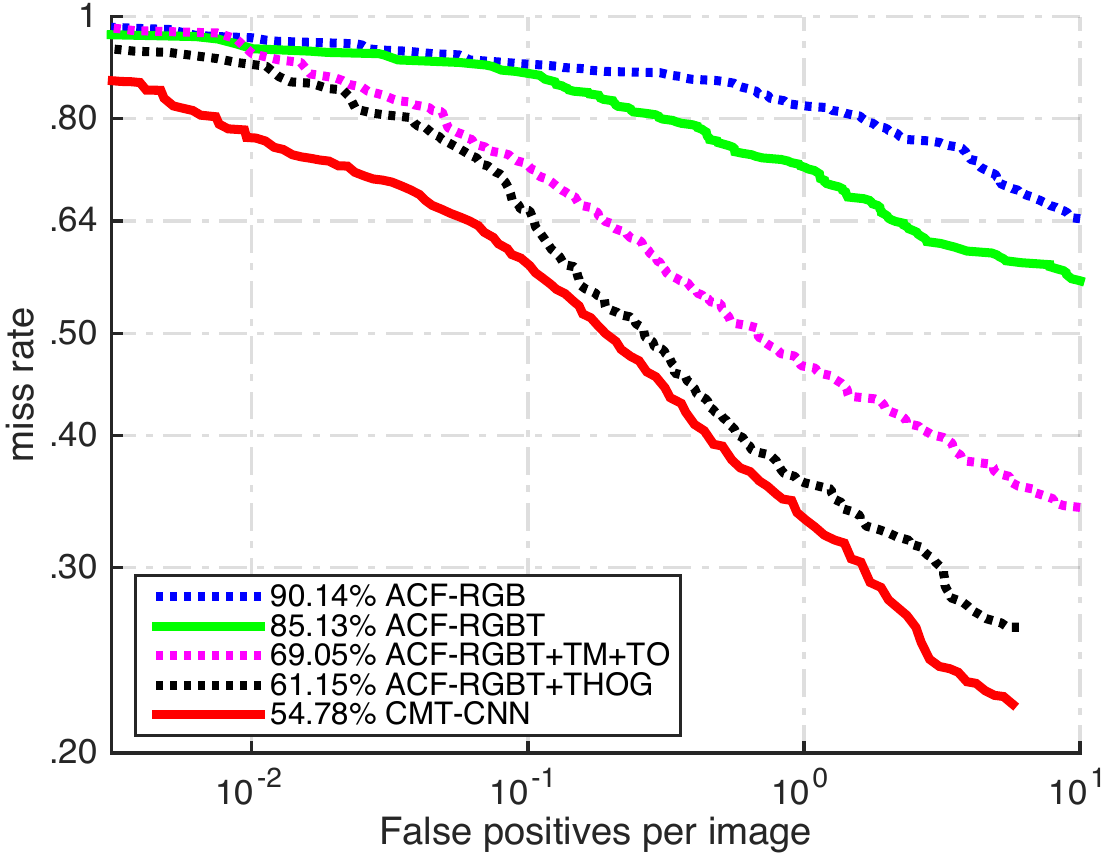}}
\caption{Quantitative evaluation results (miss rate versus false positive per image) on the KAIST multispectral dataset. 
}
\label{roc_kaist}
\end{figure*}

\begin{table}[t] 
\centering
\resizebox{0.8\linewidth}{!} {
\begin{tabular}{l|c}
\toprule 
Methods & Miss-Rate \\\hline \hline
CMT-CNN-SA & 13.76\%  \\
CMT-CNN-SA-SB(Random) & 15.89\%  \\
CMT-CNN-SA-SB(ImageNet) & 13.01\%  \\
CMT-CNN-SA-SB(RGB-KAIST) & 12.51\%  \\
CMT-CNN& 10.69\%  \\
\bottomrule
\end{tabular}
}
\caption{Comparison of different variants of our method on the Caltech-Reasonable dataset. Performance are evaluated in terms of log-average miss-rate.}
\label{tab:cmtCaltech}
\vspace{-0.5cm}
\end{table}

\subsection{Results on KAIST multispectral dataset} 

\paragraph{Analysis of proposed method.} The first series of experiments aims to demonstrate the effectiveness of the proposed
Cross-Modality Transfer CNN (CMT-CNN) framework. We evaluate the performance of our approach under four different settings: 
(i) CMT-CNN-SA. We only use Sub-Net A. The two ROI feature maps are concatenated and given as input to the convolutional fusion layer. 
This layer outputs a feature map with size 512, rather than 1024. Finally, the output is fed to the fully connected layers; 
(ii) CMT-CNN-SA-SB (ImageNet). We consider two sub-networks but initialize the convolutional layers of Sub-Net B 
from pretrained VGG16 model on ImageNet; (iii)  CMT-CNN-SA-SB (Random): Same as (ii) 
but with random initialization for Sub-Net B; (v) CMT-CNN as described in Section \ref{method}, \ie~initializing the convolutional layers 
of Sub-Net B from trained RRN. 

Table \ref{tab:cmtcmpKAIST} shows the results of our comparison. Performance is evaluated
using the log average miss-rate (MR). 
From the table it is clear that CMT-CNN significantly outperforms all its variations on all the three test sets, confirming the fact
that the proposed cross-modality framework improves the detection accuracy. We also observe that CMT-CNN provides lower MR
than CMT-CNN-SA-SB, indicating that the performance gain of CMT-CNN is not only due to an increased number of parameters.

Figure \ref{fig:reconstruction_results} depicts some examples of the reconstruction results obtained with the proposed RRN. 
For the two given test frames, the reconstructed thermal regions associated to the top nine detection windows computed with ACF are shown. 
From the figure, it is easy to observe that the proposed network is able to effectively learn a mapping from RGB data to thermal data. 
Figure~\ref{fig:detection_results} shows some qualitative results obtained with MSDN. Comparing the detection bounding boxes of CMT-CNN-SA with those of CMT-CNN, we observe that hard negative samples are correctly classified with our method. For instance, 
the foliage from the trees (Fig.~\ref{fig:detection_results}- first and second columns) is wrongly detected as pedestrian by CMT-CNN-SA. 
This confirms our intuition that leveraging information from multispectral data with our cross-modal representation transfer approach 
permits to improve the detection accuracy.
\vspace{-0.5cm}

\paragraph{Comparison with state of the art methods.} We also compare our approach with state of the art methods on the KAIST multispectral dataset. These methods include: (i) ACF-RGB~\cite{dollar2014fast}, \ie \ 
using ACF on RGB data; (ii) ACF-RGBT~\cite{hwang2015multispectral}, \ie \ using ACF on RGB-Thermal data; (iii) ACF-RGBT+TM+TO~\cite{hwang2015multispectral}, \ie \ using ACF on RGB-Thermal data with extra gradient magnitude and HOG of thermal images; (iv) ACF-RGBT+HOG~\cite{hwang2015multispectral}, \ie \ 
using ACF on RGB-Thermal data with HOG features with more gradient orientations than (iii). Results associated to these methods 
have been taken directly from the original paper \cite{hwang2015multispectral}. Similarly to baseline approaches, we also use ACF 
for generating proposals both at training and at test time.

Observing Fig.~\ref{roc_kaist}, it is clear that CMT-CNN is several points better than ACF-RGBT+HOG, the best baseline
on the KAIST dataset. Importantly, CMT-CNN only uses color images in the test phase, while ACF-RGBT+HOG exploits
both RGB and thermal data. We also observe that on the \textit{Reasonable night} setting, our approach obtains a more significant improvement  
than in the \textit{Reasonable day} experiments. This demonstrates that CMT-CNN is especially useful 
for pedestrian detection under dark illumination conditions, thus confirming our initial intuition.

\begin{table}[t] 
\centering
\resizebox{0.98\linewidth}{!} {
\begin{tabular}{l|ccccc}
\toprule 
batch size & 32 & 64 & 128 & 256 \\\hline \hline
Caltech-All & 65.97\% & 65.68\% & 65.32\% & 65.42\%\\
Caltech-Reasonable & 13.52\% & 13.01\% & 12.51\% & 12.35\%  \\
\bottomrule
\end{tabular}
}
\caption{Performance using different batch size in CMT-CNN-SA-SB (RGB-KAIST) experiments.}
\label{tab:batch-size}
\end{table}

\begin{figure*}[t]
\centering
\subfigure[ ]{\includegraphics[width=0.331\linewidth]{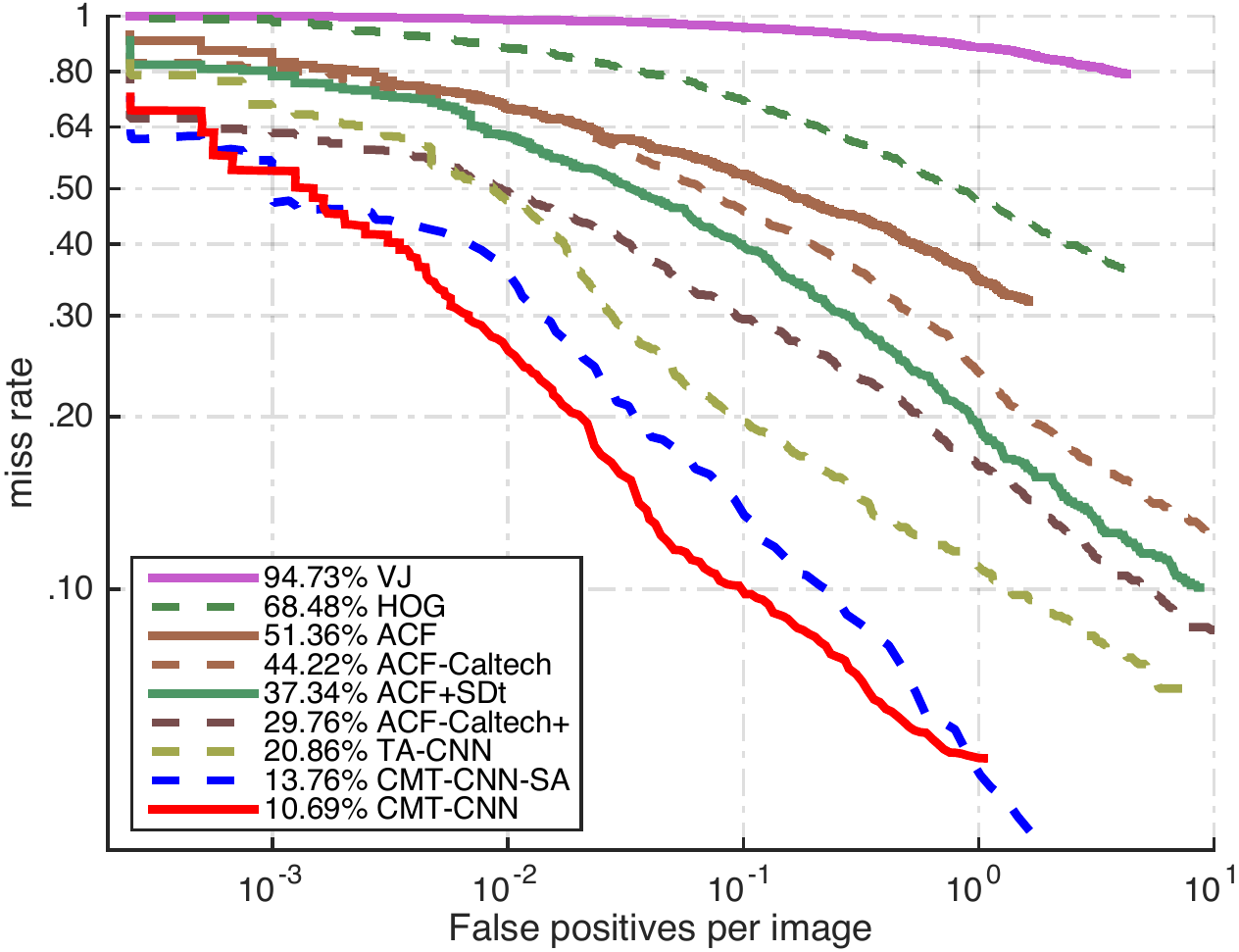}}%
\subfigure[ ]{\includegraphics[width=0.331\linewidth]{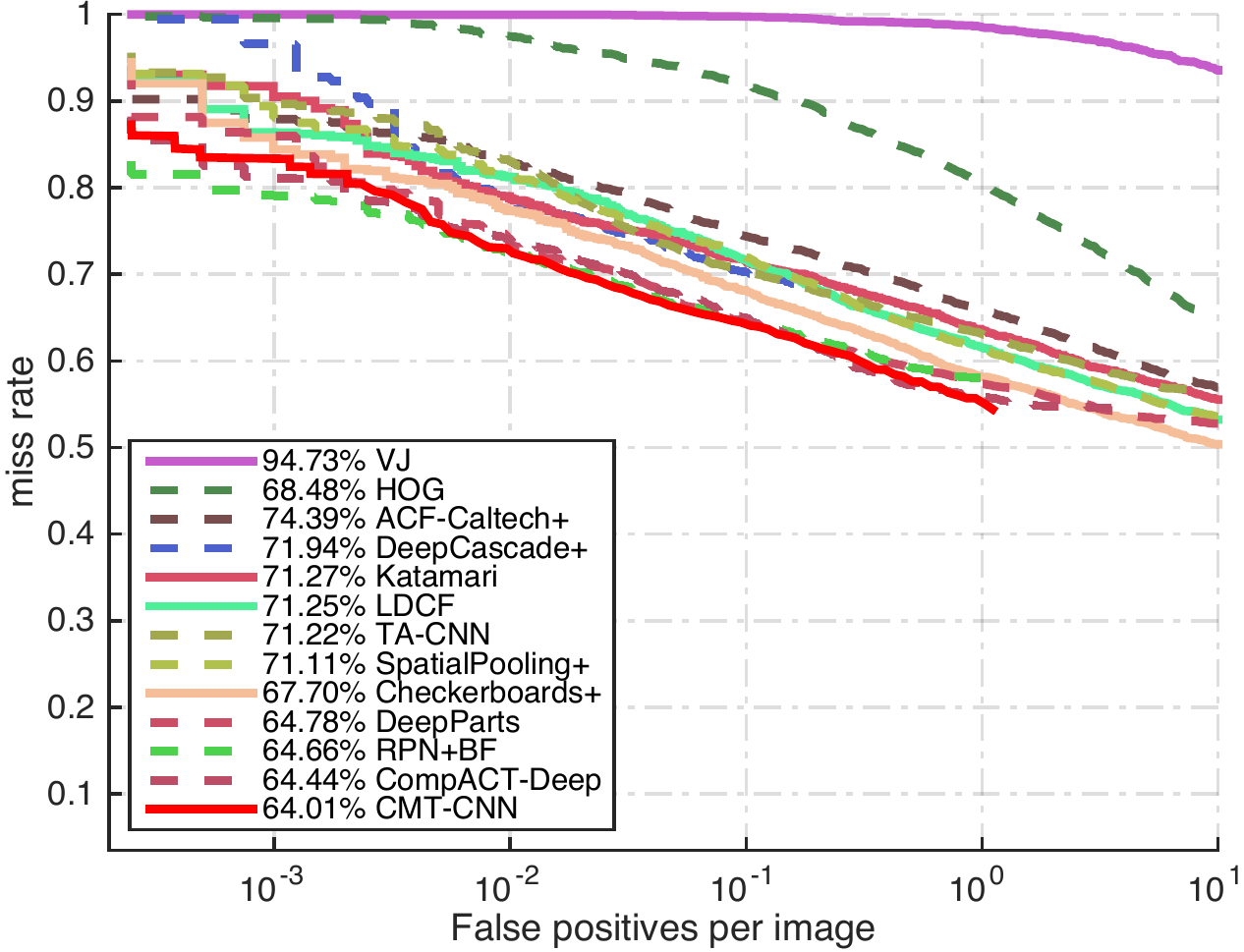}}
\subfigure[]{\includegraphics [width=0.331\linewidth]{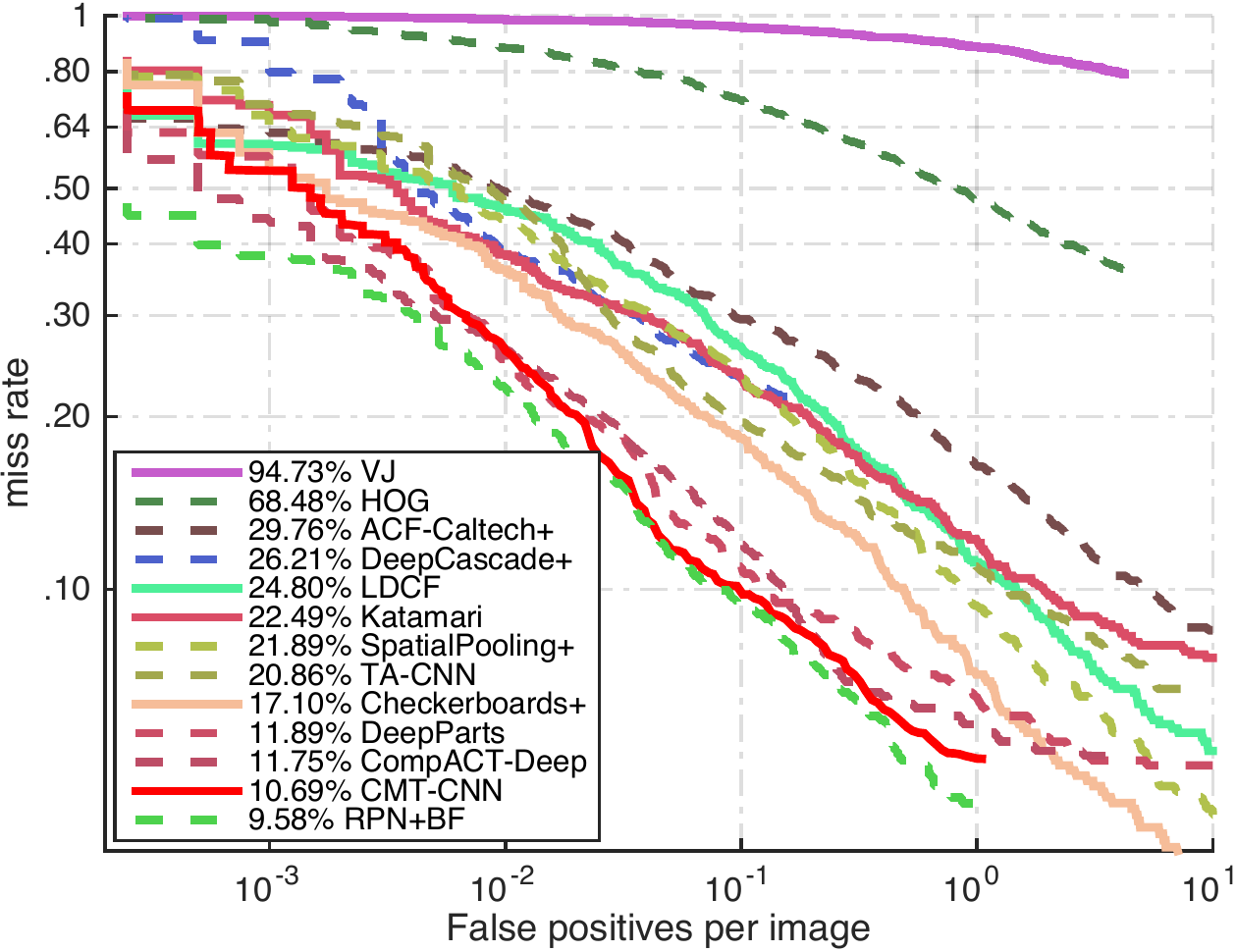}}
\caption{Quantitative evaluation results on the Caltech pedestrian dataset:  
comparison with (a) previous methods using ACF for proposals (VJ and HOG methods do not use ACF, but are kept as reference points)
(b) state of the art methods on Caltech-All (c) state of the art methods on Caltech-Reasonable. }
\label{roc_caltech}
\vspace{-0.3cm}
\end{figure*}

\begin{table}[t] 
\centering
\resizebox{1\linewidth}{!} {
\begin{tabular}{l|ccc}
\toprule 
Method & Hardware & Miss-Rate & Testing Time (s/f) \\\hline\hline
InformedHaar~\cite{zhang2014informed}  & CPU & 75.85\% & 1.59\\
SpatialPooling~\cite{paisitkriangkrai2014strengthening} & CPU & 74.04\% & 7.69 \\
LDCF~\cite{nam2014local}   & CPU & 71.25\% & 0.60 \\
CCF~\cite{yang2015convolutional}  & Titan Z GPU & 66.73\% & 13.0\\
RPN + BF~\cite{zhang2016faster} & Tesla K40 GPU & 64.66\% & 0.51 \\
CompACT-Deep~\cite{cai2015learning} & Tesla K40 GPU & 64.44\% & 0.50 \\ \hline
CMT-CNN & Tesla K40 GPU &    64.01\%    &  0.59  \\
\bottomrule
\end{tabular}
}
\caption{Comparison of different methods (log-average miss-rate vs detection time). Log-average miss-rate is evaluated on the Caltech-All. s/f represents seconds per frame.} 
\label{tab:timeCaltech}
\vspace{-0.5cm}
\end{table}
 
\subsection{Results on Caltech pedestrian dataset} 
\label{caltech_exp}
\paragraph{Analysis of CMT-CNN.} Similarly to the experiments on the KAIST dataset, we first analyze the performance of our approach when 
different initialization strategies are used for Sub-Net B. In this case we also consider another baseline CMT-CNN-SA-SB (RGB-KAIST), \ie \ we 
initialize Sub-Net B with VGG16 pretrained on ImageNet and further train it using RGB data of KAIST. 
The results of the comparison are shown in Table \ref{tab:cmtCaltech} and confirm the effectiveness of our framework. We observe that 
CMT-CNN-SA-SB (RGB-KAIST) beats CMT-CNN-SA-SB (ImageNet), showing that fine tuning CMT-CNN-SB with KAIST RGB data provides effective 
representations for improving the detection performance on Caltech. By using complementary data from the thermal modality, 
CMT-CNN further boosts its accuracy and outperforms CMT-CNN-SA-SB (RGB-KAIST). We observe that the 
improvement due to knowledge transfer on Caltech data is less pronounced than that obtained on KAIST dataset. We 
believe that this is mainly due to the fact that the frames of Caltech generally exhibit better illumination conditions than those of KAIST, 
while thermal information is especially beneficial in case of bad illumination.

To further demonstrate that the performance gain obtained with the proposed CMT-CNN 
is not simply due to ensembling different models, we consider the baseline CMT-CNN-SA-SB(RGB-KAIST) and
we train Sub-Net B with KAIST RGB images using four different mini-batch size ranging from 32 to 256.
For each experiment, the training samples are randomly shuffled. Table~\ref{tab:batch-size} shows the 
results of the four trials on Caltech-All and Caltech-Reasonable: using different 
batch size for Sub-Net B slightly affects the final performance and the best MR reported in the table
is still worse than those obtained with CMT-CNN. This
confirms the validity of our cross-modality learning approach.

We also compare the proposed CMT-CNN which 
uses ACF to generate region proposals with previous approaches also based on ACF proposals. 
Figure~\ref{roc_caltech}(a) shows the results of our comparison: our model outperforms all the baselines. Moreover, similarly to what we observed 
for KAIST experiments, CMT-CNN is more accurate than CMT-CNN-SA, confirming the advantage of our approach.
\vspace{-0.5cm}
\paragraph{Comparison with state of the art methods.} A comparison with state of the art methods is provided in Fig.~\ref{roc_caltech}(b). 
We considered Viola-Jones (VJ)~\cite{viola2004robust}, Histograms of Oriented Gradients (HOG)~\cite{dalal2005histograms}, 
DeepCascade+~\cite{angelovareal}, LDCF~\cite{nam2014local}, SCF+AlexNet~\cite{hosang2015taking}, 
Katamari~\cite{benenson2014ten}, SpatialPooling+~\cite{PaisitkriangkraiSpatially}, SCCPriors~\cite{yangexploring}, 
TA-CNN~\cite{tian2015pedestrian}, CCF and CCF+CF~\cite{yang2015convolutional}, 
Checkerboards and Checkerboards+~\cite{zhang2015filtered}, DeepParts~\cite{tian2015deep}, CompACT-Deep~\cite{cai2015learning}
and RPN+BF\cite{zhang2016faster}. 
Our approach attains a miss-rate of 10.69\% on Caltech-Reasonable, which is very competitive 
with the state of the art methods, and a miss-rate of 64.01\% on Caltech-All, which establishes a new state-of-the-art result. 
Importantly, our approach can be seen as complementary to most previous works. In fact,
we believe that our unsupervised learning of cross-modal representations can be also integrated in other CNN architectures, to improve their robustness in coping with bad illumination conditions.

In Table \ref{tab:timeCaltech} we report a comparison between our framework and recent pedestrian detection methods 
in terms of computational efficiency (times associated to previous methods are taken from the original papers). 
At test time, our network takes only 0.59 seconds to process one image, which is very competitive with previous methods.

\section{Conclusions}
\label{conclusions}
We presented a novel approach for robust pedestrian detection under adverse illumination conditions. 
Inspired by previous works on multi-scale pedestrian detection \cite{zhang2016faster}, 
a novel deep model is introduced to learn discriminative 
feature representations from raw RGB images. Differently form previous methods, 
the proposed architecture integrates a sub-network, pre-trained on pairs of RGB and thermal images, such as
to learn cross-modal feature representations. In this way,
knowledge transfer from multispectral data is achieved and accurate detection is possible even in case of challenging illumination conditions.
The effectiveness of the proposed approach is demonstrated with extensive 
experiments on publicly available benchmarks: the KAIST multispectral and the Caltech pedestrian detection datasets. 
\par While this work specifically addresses the problem of pedestrian detection, the idea behind our cross-modality learning framework can be useful in other applications (\eg, considering reconstructing depth images~\cite{xu2017multi} for RGBD object/action detection and recognition). Hence, natural directions for future research include further investigating this possibility.

{\small
\bibliographystyle{ieee}
\bibliography{egbib}
}

\end{document}